\documentclass[conference]{IEEEtran}

\IEEEoverridecommandlockouts % for identifying funding in first footnote

\usepackage{cite}
\usepackage{amsmath,amssymb,amsfonts}
\usepackage{algorithmic}
\usepackage{graphicx}
\usepackage{textcomp}
\usepackage{xcolor}
\def\BibTeX{{\rm B\kern-.05em{\sc i\kern-.025em b}\kern-.08em
    T\kern-.1667em\lower.7ex\hbox{E}\kern-.125emX}}

\usepackage[colorlinks=true, allcolors=blue]{hyperref}
\usepackage{url}
\usepackage[caption=false,font=footnotesize]{subfig} % make sure caption=false turned on to avoid interfering with IEEEtrans' handling of captions
\usepackage{algorithm,algorithmic}
\usepackage{threeparttable}

\begin{document}

\title{Practical Active Learning with Model Selection for Small Data}

%\iffalse
\author{
\IEEEauthorblockN{Maryam Pardakhti\IEEEauthorrefmark{1}}
\IEEEauthorblockA{\textit{Computer Science and Engineering; Chemical and Biomolecular Engineering}\\
\textit{University of Connecticut}\\
Storrs, USA\\
ORCID: 0000-0002-9991-0826 \\
}
\and
\IEEEauthorblockN{Nila Mandal\IEEEauthorrefmark{1}} 
\IEEEauthorblockA{\textit{Computer Science and Engineering}\\
\textit{University of Connecticut}\\
Storrs, USA\\
ORCID: 0000-0003-3088-4655 \\
}
\and
\IEEEauthorblockN{Anson W. K. Ma}
\IEEEauthorblockA{\textit{Chemical and Biomolecular Engineering}\\
\textit{University of Connecticut}\\
Storrs, USA\\
ORCID: 0000-0002-2865-5776 \\
}
\and
\IEEEauthorblockN{Qian Yang\IEEEauthorrefmark{2}}
\IEEEauthorblockA{\textit{Computer Science and Engineering}\\
\textit{University of Connecticut}\\
Storrs, USA\\
ORCID: 0000-0001-5519-1092 \\
}
\thanks{This work is supported by the USDA National Institute of Food and Agriculture, AFRI project 2019-06721. The authors also acknowledge support from the US Department of Education (GAANN Fellowship to N.M. under award no. P200A180092) and the University of Connecticut Research Excellence Program.
\newline\IEEEauthorrefmark{1} Equal contribution.
\newline\IEEEauthorrefmark{2} Corresponding author.
\newline
\newline\copyright 2021 IEEE. Personal use of this material is permitted. Permission from IEEE must be obtained for all other uses, in any current or future media, including reprinting/republishing this material for advertising or promotional purposes, creating new collective works, for resale or redistribution to servers or lists, or reuse of any copyrighted component of this work in other works.
\newline DOI 10.1109/ICMLA52953.2021.00263
}
}
%\fi
%\author{Anonymous Authors}

\maketitle

% As a general rule, do not put math, special symbols or citations in the abstract
\begin{abstract}

Active learning is of great interest for many practical applications, especially in industry and the physical sciences, where there is a strong need to minimize the number of costly experiments necessary to train predictive models. However, there remain significant challenges for the adoption of active learning methods in many practical applications. One important challenge is that many methods assume a fixed model, where model hyperparameters are chosen \textit{a priori}. In practice, it is rarely true that a good model will be known in advance. Existing methods for active learning with model selection typically depend on a medium-sized labeling budget. In this work, we focus on the case of having a very small labeling budget, on the order of a few dozen data points, and develop a simple and fast method for practical active learning with model selection. Our method is based on an underlying pool-based active learner for binary classification using support vector classification with a radial basis function kernel. First we show empirically that our method is able to find hyperparameters that lead to the best performance compared to an oracle model on less separable, difficult to classify datasets, and reasonable performance on datasets that are more separable and easier to classify. Then, we demonstrate that it is possible to refine our model selection method using a weighted approach to trade-off between achieving optimal performance on datasets that are easy to classify, versus datasets that are difficult to classify, which can be tuned based on prior domain knowledge about the dataset.

\end{abstract}


% For peer review papers, you can put extra information on the cover
% page as needed:
% \ifCLASSOPTIONpeerreview
% \begin{center} \bfseries EDICS Category: 3-BBND \end{center}
% \fi
%
% For peerreview papers, this IEEEtran command inserts a page break and
% creates the second title. It will be ignored for other modes.
\IEEEpeerreviewmaketitle

\section{Introduction}

In many applications in science and industry, useful labeled data are very expensive or difficult to collect, while unlabeled data are extensively available. A promising and increasingly popular approach in these cases is to use active learning, which allows practitioners to choose the most informative data points to label given a limited budget\cite{Cohn1996,Settles2011,Settles2012}. However, it is difficult to do model selection with such limited data, so methods for active learning usually use a fixed model. Since it is difficult to know \textit{a priori} what model works best for a given problem and dataset, this can often lead to suboptimal performance, and represents a significant challenge for practical applications of active learning in scientific and industrial settings. In this work, we consider active learning for binary classification using support vector machines with a radial basis function kernel, and present a method for active learning with model selection that is designed to work for very small labeling budgets of up to a few dozen datapoints. This precludes the use of many existing methods for model selection currently in the literature that depend on medium-sized labeling budgets of up to hundreds of datapoints in order to achieve improved performance over fixed models. We also propose a variation of this method that improves performance on datasets that are more separable and thus easier to classify, and discuss how it leads to a trade-off in performance on datasets that are less separable and thus more difficult to classify. When prior information is known about the expected separability of the given dataset, we can use it to choose which of the two methods to apply.

We first present a review of related works in Section~\ref{sec:related}. In Section~\ref{sec:methods}, we introduce our two proposed algorithms for active learning with model selection. In Section~\ref{sec:experiments} we describe our experiments and in Section~\ref{sec:results} we present the results of evaluating the proposed algorithms on several benchmark datasets compared to different baselines. We also examine the circumstances under which each of our two model selection methods is likely to be more or less effective. Finally in Section~\ref{sec:conclusions}, we discuss key takeaways and conclusions.

\section{Related work}
\label{sec:related}

Active learning has received significant attention in machine learning for various types of problems, from regression to classification to reinforcement learning \cite{Belharbi2021,Menard2021,Settles2011}. In active learning, the key idea is to sample datapoints efficiently to optimize model performance. Pool-based sampling is one widely used approach, where the acquisition algorithm queries for new samples to label from a pool of unlabeled data. Pool-based active learning using uncertainty sampling and support vector machine learners is a popular strategy among these approaches\cite{Campbell2000, Schohn2000, Tong2001, Kremer2014}.

While active learning has shown promising results, many obstacles for its practical application remain which undermines its effectiveness especially in scientific and industrial applications. Challenges include the need for \textit{a priori} knowledge in order to choose the best active learning method and model parameters, and the difficulty of handling complex data distributions that are either not known beforehand and/or make active learning ineffective (i.e., skewed or imbalanced data, disjunctive classes) \cite{Attenberg2011,Lowell2019}. Our proposed method for model selection attempts to address the first challenge in the limited setting of pool-based active learning with uncertainty sampling using support vector classifiers.

So far in the literature, most active learning studies have focused on using a fixed active learning method and model. Model selection as part of the active learning process is challenging to implement, and only a small number of approaches have been studied. Sugiyama and Reubens showed that in active learning for regression problems instead of using a fixed model, training an ensemble of models leads to better performance\cite{Sugiyama2008}. In another study, Ali et al.\cite{Ali2014} introduced a novel approach for model selection in which the labeled samples are divided into separate training and validation sets in order to train the models and select the best model, respectively. This method is effective as it creates a validation set to select the best learner but is expensive to implement. Both approaches also require more data to be labeled to enable model selection, which is not practical for very small labeling budgets\cite{Ali2014}.

Our proposed method for data-efficient model selection relies on reusing the biased set of datapoints labeled during active learning. The reusability of actively sampled data remains a challenging area of research\cite{Baldridge2004, Tomanek2007, Tomanek2011, Hu2016}. Empirical studies suggest that sample reusability for a given selector-consumer combination, where the selector is the algorithm that generates the actively sampled dataset and the consumer is the algorithm used to train the final model, is difficult to generalize across learning problems\cite{Tomanek2011}. It has also been suggested empirically that support vector machine learners may be preferable to other classifiers such as k-nearest neighbors and na\"{i}ve Bayes as the selector when the consumer is not known beforehand\cite{Hu2016}.

\section{Methods}
\label{sec:methods}

In this work, we develop two methods for binary classification with model selection in cases where the labeling budget is very small, on the order of a few dozen datapoints. A key difficulty when the budget is so small is that any method that requires a separate validation set would be infeasible. A small validation set will not be able to effectively distinguish between the performance of many different models. Instead, our first method relies on reusing data actively sampled by a fixed underlying active learning model. We implement model selection via leave-one-out cross validation (LOOCV) on this data. Typically, since the actively sampled data is biased, it is understood that LOOCV will lead to biased estimates of the generalization error. However, we will show empirically that when a dataset is inherently difficult to classify, LOOCV is able to \textit{order} the models such that a higher performing model can often be found. When a dataset is easy to classify, we note that LOOCV underestimates the accuracy of the model corresponding to the active learner, and use this intuition to propose a second method that implements a correction. This correction improves model selection performance for ``easy-to-classify" datasets, at the expense of possibly degrading model selection performance on ``difficult-to-classify" datasets. In this work, we also develop a methodology for characterizing whether a dataset is likely to be easy or difficult to classify; while this characterization would not be known \textit{a priori} for a new problem, we can use it to illustrate the trade-off in performance between different model selection methods with respect to different types of datasets in our benchmark experiments. This will enable us to build intuition about which method to use given prior domain knowledge about a dataset, which we will discuss further in Section~\ref{sec:results}.

We note that importantly, in both of our proposed methods model selection takes place after active learning has completed using a fixed model, rather than attempting to utilize information about model performance for different models when acquiring actively sampled datapoints. This prevents degradation of the performance of the active learner, since any estimates of model performance will have high variance especially in the beginning of active learning when the labeled dataset is very small.

In this work, our underlying active learner uses pool-based active learning with a support vector classifier (SVC) with radial basis function (RBF) kernel. The optimization problem being solved for the SVC can be written in primal form as 
\begin{align*}
    \min_{w,b,\xi} \hspace{2mm}&\hspace{2mm} \frac{1}{2}w^Tw + C\sum_{i=1}^n \xi_i \\
    \text{subject to} \hspace{2mm}&\hspace{2mm} y_i \left(w^T\phi(x_i)+b\right) \geq 1-\xi_i \\
    &\hspace{2mm} \xi_i \geq 0 , \hspace{5mm} i=1,...,n
\end{align*}
A radial basis function kernel given by
\begin{align*}
    K(x_i,x_j) & = \exp{\left(-\gamma\|x_i-x_j\|^2\right)} \\
    & = \langle \phi(x_i), \phi(x_j) \rangle
\end{align*}
is used to learn the nonlinear decision boundary. Our model selection problem is to choose the optimal hyperparameters $C$ and $\gamma$ for the SVC. Hyperparameter $C$ is a regularization parameter that represents the trade-off between maximizing the margin and minimizing errors (misclassifications and datapoints within the margin) in soft-margin SVC. Small values of $C$ correspond to larger margins and thus a simpler decision function that is less likely to overfit. The intuitive explanation of hyperparameter $\gamma$ is that it is inversely proportional to the spread of the RBF kernel, which encodes the radius of influence of each training datapoint. Small values of $\gamma$ correspond to a large spread for the RBF kernel, and thus a simpler decision function that is less likely to overfit; conversely, for very large values of $\gamma$ the decision function tends to circle around every datapoint and leads to overfitting. Commonly used values of SVM hyperparameters are $C=1$ and $\gamma=\frac{1}{n}$, where $n$ is the number of features, although when the dataset is sufficiently large the optimal $C$ and $\gamma$ are typically chosen via k-fold cross validation. Unfortunately, this is nontrivial when combined with active learning because we need to first choose $C$ and $\gamma$ before we can acquire new datapoints for the training set.

Our first proposed algorithm for active learning with model selection is given in Algorithm~\ref{alg:palms}. The inputs to our algorithm are the following:
\begin{itemize}
    \item initial labeled dataset, $\mathcal{D}_{\text{init}}$
    \item unlabeled data pool, $\mathcal{D}_{\text{pool}}$
    \item active learning budget, $B$
    \item fixed model $M_{\text{default}} = (C_{\text{default}}, \gamma_{\text{default}})$ for active learning
    \item set of models $\{M\}_{i=1}^m$ (including $M_{\text{default}}$) for model selection
\end{itemize}
First, we train an SVC with fixed hyperparameters $(C_{\text{default}}, \gamma_{\text{default}})$ on an initial set of labeled training data $\mathcal{D}_{\text{init}}$. Then, the active learner queries the label for the sample in the data pool with the shortest distance to the currently trained model's decision boundary, as described in~\cite{Campbell2000,Schohn2000,Tong2001}. The acquisition function we use is
\begin{equation}
    x^* = \text{argmin}_{x\in\mathcal{P}} \hspace{2mm} d(x,\theta)
\label{eqn:acquisition}
\end{equation}
where $d(x,\theta)$ is the distance of datapoint $x\in\mathbb{R}^n$ from the currently trained model's decision boundary, denoted by $\theta$. After labeling this point, it is added to the training data and the model is retrained with this updated training set. This process continues until we have reached the active learning budget. Then, using all of the labeled data sampled from the active learning process, we use LOOCV to select the model with the highest accuracy among the set of models $\{M\}_{i=1}^m$. If two or more models having the same LOOCV error are tied for the best model, ties are broken by first checking which model has the smaller $\gamma$ parameter, and if they are equal, which has the smaller $C$ parameter. In practice, due to the small labeling budget, these tiebreaks are often necessary. Note that smaller $\gamma$ and $C$ correspond to simpler decision boundaries that are less likely to overfit. Due to the nature of the RBF kernel, we observe empirically that generalization error is typically lower when we break ties using the $\gamma$ parameter first rather than the $C$ parameter first. Larger values of the $\gamma$ parameter tend to lead to decision boundaries that encircle datapoints and may create disjoint regions in feature space with the same class. 

Our method can be implemented on a very small labeling budget which is often the case for practical problems in industry, so we call our method practical active learning with model selection (PALMS).

\begin{algorithm}
 \caption{Practical Active Learning with Model Selection (PALMS)}
 \begin{algorithmic}[1]
 \renewcommand{\algorithmicrequire}{\textbf{Input:}}
 \renewcommand{\algorithmicensure}{\textbf{Output:}}
 \REQUIRE Initial labeled dataset $\mathcal{D}_{\text{init}}$, unlabeled dataset $\mathcal{D}_{\text{pool}}$, active learning budget $B$, fixed model $M_{\text{default}}$ for active learning, set of models $\{M\}_{i=1}^m$ (including $M_{\text{default}}$) for model selection
 \ENSURE  Best-performing model $M^* \in \{M\}_{i=1}^m$ trained on labeled dataset from active learning. 
 \\ \textit{Initialization} : Set training set $\mathcal{T} \leftarrow \mathcal{D}_{\text{init}}$; set data pool $\mathcal{P} \leftarrow \mathcal{D}_{\text{pool}}$;
  \FOR{$i=1$ to $B$}
  \STATE Train $M_{\text{default}}$ on $\mathcal{T}$;
  \STATE Compute the next query sample $x^* \in \mathcal{P}$ according to Eqn.~\ref{eqn:acquisition};
  \STATE Request label $y^*$ for $x^*$;
  \STATE Set $\mathcal{T} \leftarrow \mathcal{T} \cup (x^*,y^*)$;
  \STATE Set $\mathcal{P} \leftarrow \mathcal{P} \setminus (x^*,y^*)$;
  \ENDFOR
 \\ Choose the best model $M^* \in \{M\}_{i=1}^m$ using LOOCV on $\mathcal{T}$;
 \\ Train $M^*$ on $\mathcal{T}$;
 \end{algorithmic} 
 \label{alg:palms}
\end{algorithm}

\begin{figure}[t]
    \includegraphics[width=\columnwidth]{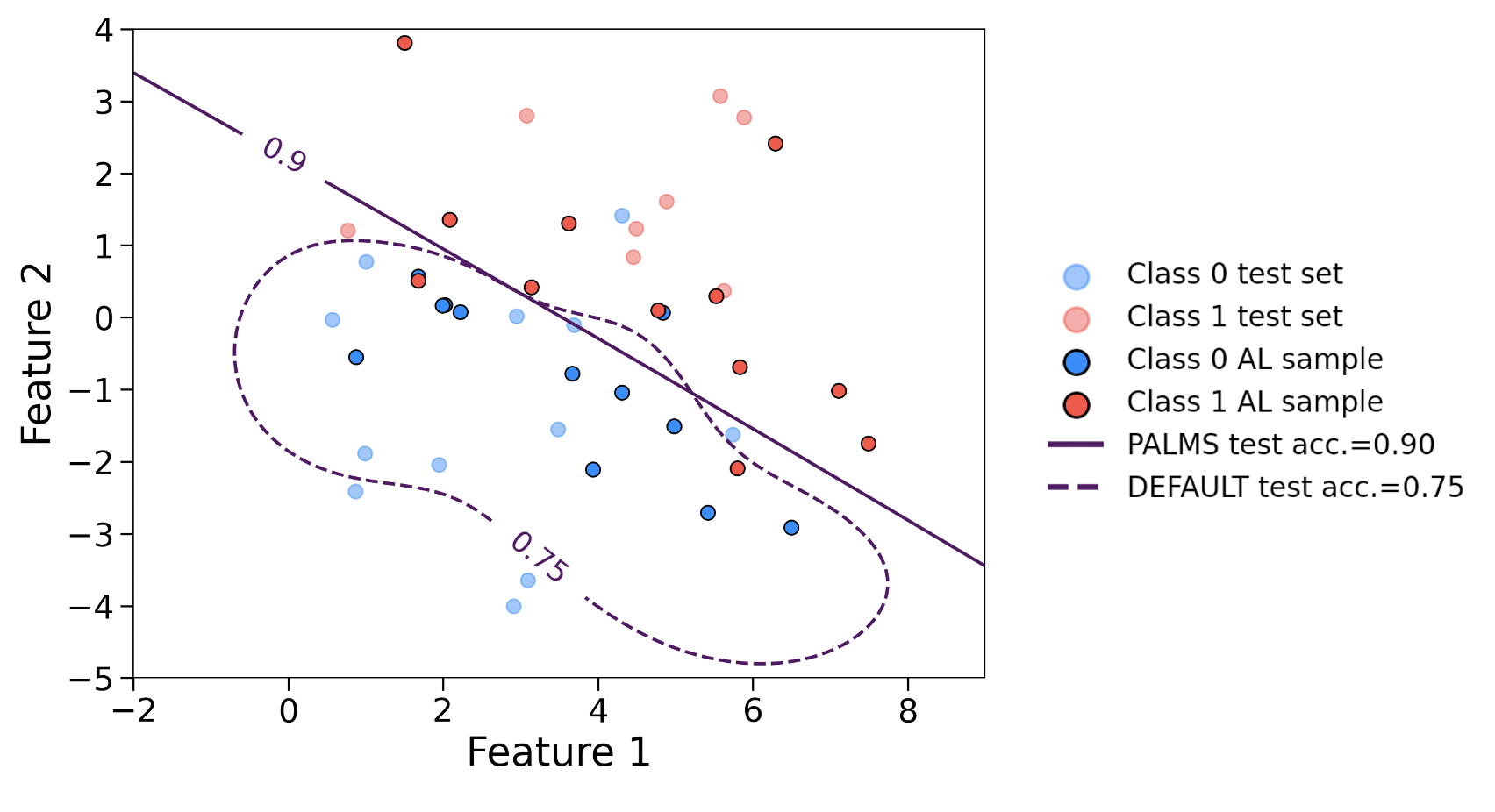}
    \caption{Example of PALMS on a toy dataset with 24 total labeled datapoints (AL sample), consisting of 4 initial datapoints and 20 requested labels from active learning. The fixed DEFAULT model generates the dotted decision boundary, which achieves a test accuracy of 0.75. After model selection using PALMS, which performs LOOCV on the AL sample, the solid decision boundary is generated, achieving a test accuracy of 0.9.}
    \label{fig:boundaries}
\end{figure}

The importance of model selection is illustrated on a toy dataset with two features per datapoint in Figure~\ref{fig:boundaries}. Here the dotted line shows the decision boundary of the SVC model trained using active learning with fixed hyperparameters $C_{\text{default}}=1$ and $\gamma_{\text{default}} = 1/2$ (DEFAULT). The solid line is a different model $(C^*, \gamma^*)$ identified using PALMS. The PALMS model achieves a test accuracy of 0.9, which outperforms the DEFAULT model with a test accuracy of 0.75. Visually, we can see that PALMS finds a simpler model than DEFAULT that is less likely to overfit, leading to higher performance on the test set.

While PALMS is a simple method for model selection, it does not take into account that LOOCV generally underestimates the generalization error for the underlying fixed model used to actively sample the training data. Intuitively, this is due to the fact that the actively sampled dataset is by construction biased to be closer to the decision boundary learned using the fixed model. A reasonably separable dataset (e.g., one that is easier to classify) will have many datapoints further away from this decision boundary; the actively sampled dataset undersamples those regions, and so the LOOCV error will be worse than the true generalization error. Since PALMS uses LOOCV for all models, it will achieve suboptimal performance when the true best model is the fixed model. Therefore, we develop a second algorithm, PALMS with weight-corrected fixed model (PALMS-fwc) to apply a correction to the LOOCV error of only the fixed model when doing model selection. We use a simple heuristic to give actively sampled datapoints further away from the decision boundary a higher weight when computing the LOOCV error for the fixed model. All datapoints on each side of the decision boundary are weighted by some constant $w > 1$ if they are more than the median distance away from the decision boundary. Note that PALMS-fwc reduces to PALMS when $w=1$.

The full algorithm for PALMS-fwc is given in Algorithm~\ref{alg:palmsfwc}. We compute the cutoff distance for which datapoints will be weighted using the following equations:
\begin{equation}
    d_i = \text{median}(\{ d(x,\theta) \text{ for } (x,y)\in\mathcal{T} \text{ and } \hat{y}=i\})
\label{eqn:mediancutoff}
\end{equation}
where $\mathcal{T}$ is the current training set, $\hat{y}$ is the predicted label for $x$ given by decision boundary $\theta$, and $i\in [0,1]$ represents the two class labels. Then for a given constant weight $w>1$, the training datapoints $(x_j, y_j)$ are weighted by $w_j$ as follows:
\begin{equation}
    w_j = \begin{cases}
      w & \text{if } d(x_j,\theta) \geq d_0 \text{ and } \hat{y}_j=0 \\
      w & \text{if } d(x_j,\theta) \geq d_1 \text{ and } \hat{y}_j=1 \\
      1 & \text{otherwise}
    \end{cases}  
\label{eqn:sampleweights}
\end{equation}
Note that weighted datapoints are only determined by their distance from the decision boundary; incorrectly classified datapoints far away from the decision boundary will also be weighted. When the data is truly more separable (i.e., easier to classify), a higher weight $w$ will improve the LOOCV estimate for the fixed model; conversely, this is not true when the data are not as separable. Thus, the weighting value used will impose a tradeoff between these two cases, which is usually not possible to know \textit{a priori}.

\begin{algorithm}
 \caption{PALMS with Weight-Corrected Fixed Model (PALMS-fwc)}
 \begin{algorithmic}[1]
 \renewcommand{\algorithmicrequire}{\textbf{Input:}}
 \renewcommand{\algorithmicensure}{\textbf{Output:}}
 \REQUIRE Initial labeled dataset $\mathcal{D}_{\text{init}}$, unlabeled dataset $\mathcal{D}_{\text{pool}}$, active learning budget $B$, fixed model $M_{\text{default}}$ for active learning, set of models $\{M\}_{i=1}^m$ (including $M_{\text{default}}$) for model selection
 \ENSURE  Best-performing model $M^* \in \{M\}_{i=1}^m$ trained on labeled dataset from active learning. 
 \\ \textit{Initialization} : Set training set $\mathcal{T} \leftarrow \mathcal{D}_{\text{init}}$; set data pool $\mathcal{P} \leftarrow \mathcal{D}_{\text{pool}}$;
  \FOR{$i=1$ to $B$}
  \STATE Train $M_{\text{default}}$ on $\mathcal{T}$;
  \STATE Compute the next query sample $x^* \in \mathcal{P}$ according to Eqn.~\ref{eqn:acquisition};
  \STATE Request label $y^*$ for $x^*$;
  \STATE Set $\mathcal{T} \leftarrow \mathcal{T} \cup (x^*,y^*)$;
  \STATE Set $\mathcal{P} \leftarrow \mathcal{P} \setminus (x^*,y^*)$;
  \ENDFOR
 \\ Train $M_{\text{default}}$ on $\mathcal{T}$;
 \\ Compute cutoff distances $d_0$ and $d_1$ according to Eqn~\ref{eqn:mediancutoff};
  \FOR{$j = 1,\dots,|\mathcal{T}|$}
  \STATE Set weight $w_j$ for datapoint $x_j \in\mathcal{T}$ using $d_0$, $d_1$ according to Eqn~\ref{eqn:sampleweights};
  \ENDFOR
 \\ Set $\mathcal{W} \leftarrow \{w_j\}_{j=1}^{|\mathcal{T}|}$;
 \\ Choose the best model $M^* \in \{M\}_{i=1}^m$ on $\mathcal{T}$ using LOOCV $\forall M_i \in \{M\}_{i=1}^m \setminus M_{\text{default}}$, and weighted LOOCV with weights $\mathcal{W}$ for $M_{\text{default}}$;
 \\ Train $M^*$ on $\mathcal{T}$;
 \end{algorithmic}
 \label{alg:palmsfwc}
\end{algorithm}

\section{Experiments}
\label{sec:experiments}

We compare the performance of \textbf{PALMS}, our proposed algorithm for active learning with model selection, and \textbf{PALMS-fwc}, which adds a weight correction for the underlying fixed model as described above, to three different baseline methods:
\begin{itemize}
    \item \textbf{RANDOM} First, we use random independent, identically distributed (IID) sampling with a support vector classifier (SVC) using a radial basis function (RBF) kernel. LOOCV is used to choose the optimal model parameters $C^{\text{rand}}$ and $\gamma^{\text{rand}}$ from all combinations $(C,\gamma)$ in consideration from among the set of models $\{M\}_{i=1}^m$. This is a na\"{i}ve baseline to compare against active learning with model selection.
    \item \textbf{DEFAULT} Here we use only pool-based active learning with SVC using a RBF kernel with fixed model parameters $C=1$ and $\gamma= \frac{1}{n}$. This is the same as PALMS without the model selection step.
    \item \textbf{ORACLE} This is pool-based active learning with SVC using a RBF kernel with the best possible model parameters $M_{\text{oracle}} = (C^{\text{oracle}}, \gamma^{\text{oracle}})$ out of all model parameter combinations $(C,\gamma)$ in consideration from among the set of models $\{M\}_{i=1}^m$. The best model is determined by its test accuracy. The goal of model selection algorithms is to find a model that achieves as close as possible to ORACLE performance.
\end{itemize}
The range of values for $C$ and $\gamma$ considered in this work are $[0.01,1,100,10^4]$ and $\frac{1}{n}[10^{-4}, 0.01, 1, 100, 10^4]$, respectively.

We evaluate these methods on several benchmark datasets for binary classification, described in Table~\ref{tab:datasets}. For each dataset, first a subset of 50 labeled datapoints per class were randomly selected and set aside, to create a balanced test data set of 100 points total. Then, an initial sample of 4 datapoints is randomly selected; to generate this initial sample we use stratified sampling so that there are 2 datapoints from each class in order to make leave one out cross validation possible. While this would not be possible in a real use case scenario, we initialize our experiments this way in order to enable comparison of models trained with the same number of total labeled datapoints; in reality we would need to randomly sample the unlabeled dataset until datapoints from both classes are found or the data budget is exhausted. Finally, for each of our benchmark datasets, the remainder of the data that has not been assigned to the test set or the initial labeled dataset is considered to be the pool of unlabeled data. We repeat each experiment for 50 trials, and record the accuracy on the test set for an active learning labeling budget (or random sampling budget in the case of RANDOM) of up to 55 datapoints, excluding the 4 datapoints selected for initialization.

For each dataset, we also compute what we call the \textit{limited set} from the sampled test set. This is a subset of our test set consisting of points whose nearest neighbors are a mix of classes, which correspond to uncertain regions in feature space. We use this to characterize how difficult we should expect it to be to classify a given dataset. Datasets with large limited sets are likely to be difficult to classify in comparison to datasets with small limited sets. We use two parameters to define the limited set: $k$, the number of nearest neighbors to consider, and a cutoff threshold $\rho$ to determine how mixed the neighborhood must be. The limited set varies as $k$ and $\rho$ change. In this work, we choose $k=20$ and $\rho=0.3$. This pair of parameters means that for a point to be part of the limited set, more than $\rho k = 6$ of its $k=20$ nearest neighbors are in the minority class. 

\begin{figure}[t]
    \includegraphics[width=\columnwidth]{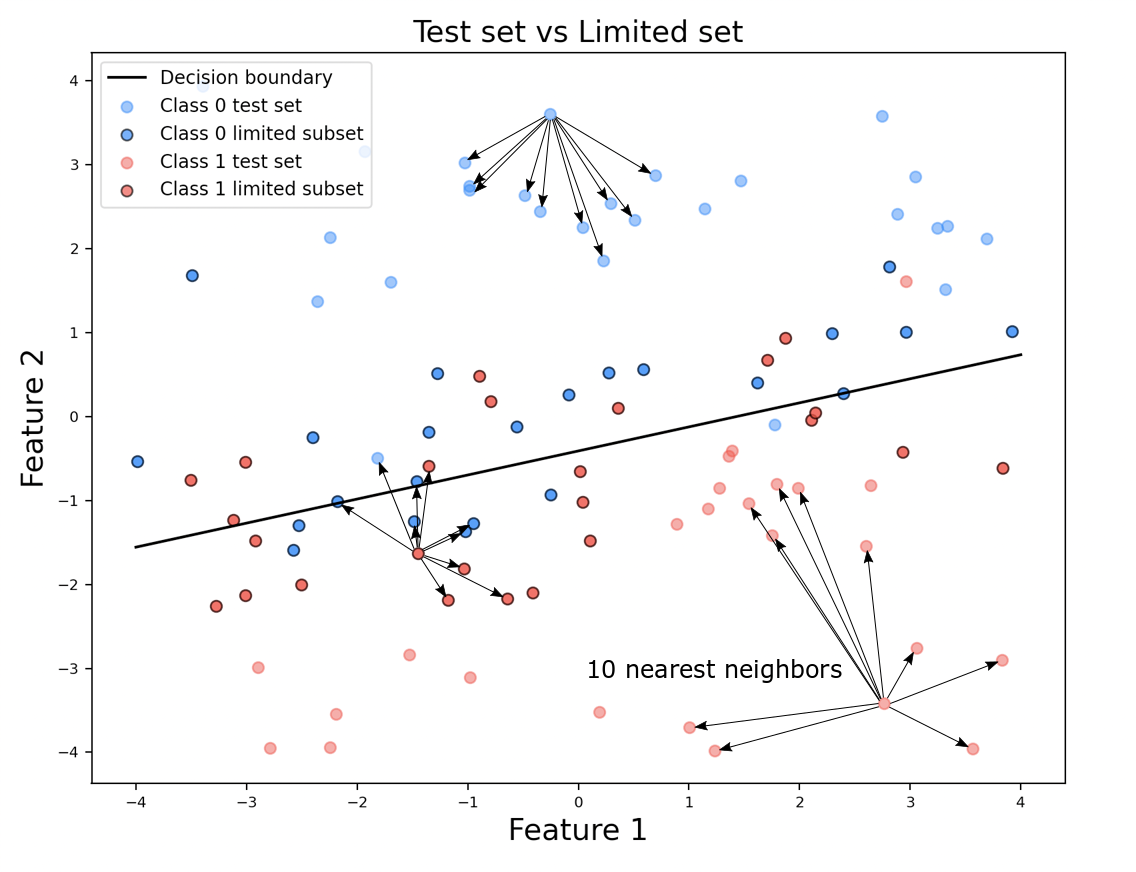}
    \caption{The \textit{limited set} is the subset of the \textit{test set}, consisting of points whose nearest neighbors are a mix of classes; datapoints in the limited set correspond to uncertain regions in feature space. Two parameters are used to define which datapoints belong to the limited set: \textit{k}, the number of nearest neighbors to consider, and a cutoff threshold $\rho$ to determine how mixed the neighborhood must be. In this figure, we illustrate the limited set for $k=10$ and $\rho=0.3$: a datapoint whose 10 nearest neighbors have more than 3 datapoints in the ``minority" class is considered part of the limited subset (darker colors). In Table~\ref{tab:datasets}, we use $k=20$ and $\rho=0.3$. Note this limited set is used only as a tool to characterize each dataset; it is not available in a real use case scenario for active learning because we would not have a labeled test set. We show that the performance of PALMS and PALMS-fwc varies with respect to the size of the limited set. }
    \label{fig:limitedset}
\end{figure}

We illustrate our method for identifying the limited set for the sample case of $k=10$ and $\rho=0.3$ in Figure~\ref{fig:limitedset}. Three example points are depicted, one in the limited set (darker colors) and two not in the limited set (lighter colors). The datapoint in the limited set has four neighbors from the opposite class. Datapoints in the limited set are usually expected to be near the decision boundary where greater uncertainty is expected. However, they may also be located in other regions in feature space away from the decision boundary learned by a given model. We note that the limited set is used only as a tool to characterize how easy or difficult we expect classification to be on each dataset; it is not available in a real use case scenario for active learning because we would not have a labeled test set. We will show empirically that the performance of PALMS and PALMS-fwc varies with respect to the size of the limited set.

\begin{table}
    \begin{threeparttable}[t]
    \caption{Datasets used in this work.\tnote{$\mathrm{a}$}}
    \label{tab:datasets}
    \centering
    \begin{tabular}{l|c|c|c|c|c}
    Dataset & features & class 0 & class 1 & total & \%limited \\ \hline
    Tictactoe \cite{Tang2019} & 9 & 332 & 626 & 958 & 92\% \\
    Solarflare\cite{Dua:2019} & 25 & 356 & 171 & 527 & 87\%\\
    KRvsKP \cite{Tang2019} & 36 & 1669 & 1527 & 3196 & 84\% \\
    Pima\cite{Dua:2019}& 8 & 500 & 268 & 768 & 71\% \\
    Spinal\cite{UCI-Kaggle-Spine} & 12 & 210 & 100 & 310 & 67\% \\
    IBN\_Sina \cite{ibnsina} & 92 & 12880 & 7842 & 20722 & 49\% \\
    Heart\cite{Dua:2019} & 13 & 138 & 164 & 302 & 48\% \\
    Phoneme \cite{Tang2019} & 5 & 3789 & 1560 & 5349 & 48\% \\
    Ionosphere\cite{Dua:2019} & 34 & 125 & 225 & 350 & 39\% \\
    Thyroid \cite{Tang2019} & 5 & 65 & 150 & 215 & 20\% \\
    \end{tabular}
    \begin{tablenotes}
        \item [$\mathrm{a}$]Columns correspond to the number of features, number of datapoints in each class, the total number of datapoints, and the average proportion of points that are in the limited subset out of a randomly sampled balanced test set. Generally, a lower proportion of points in the limited set corresponds to an easier binary classification problem for which higher accuracies can be achieved (see Figure~\ref{fig:performance}).
    \end{tablenotes}
    \end{threeparttable}
\end{table}

\section{Results}
\label{sec:results}

\begin{figure*}[t]
    \centering
    \includegraphics[width=6.5in]{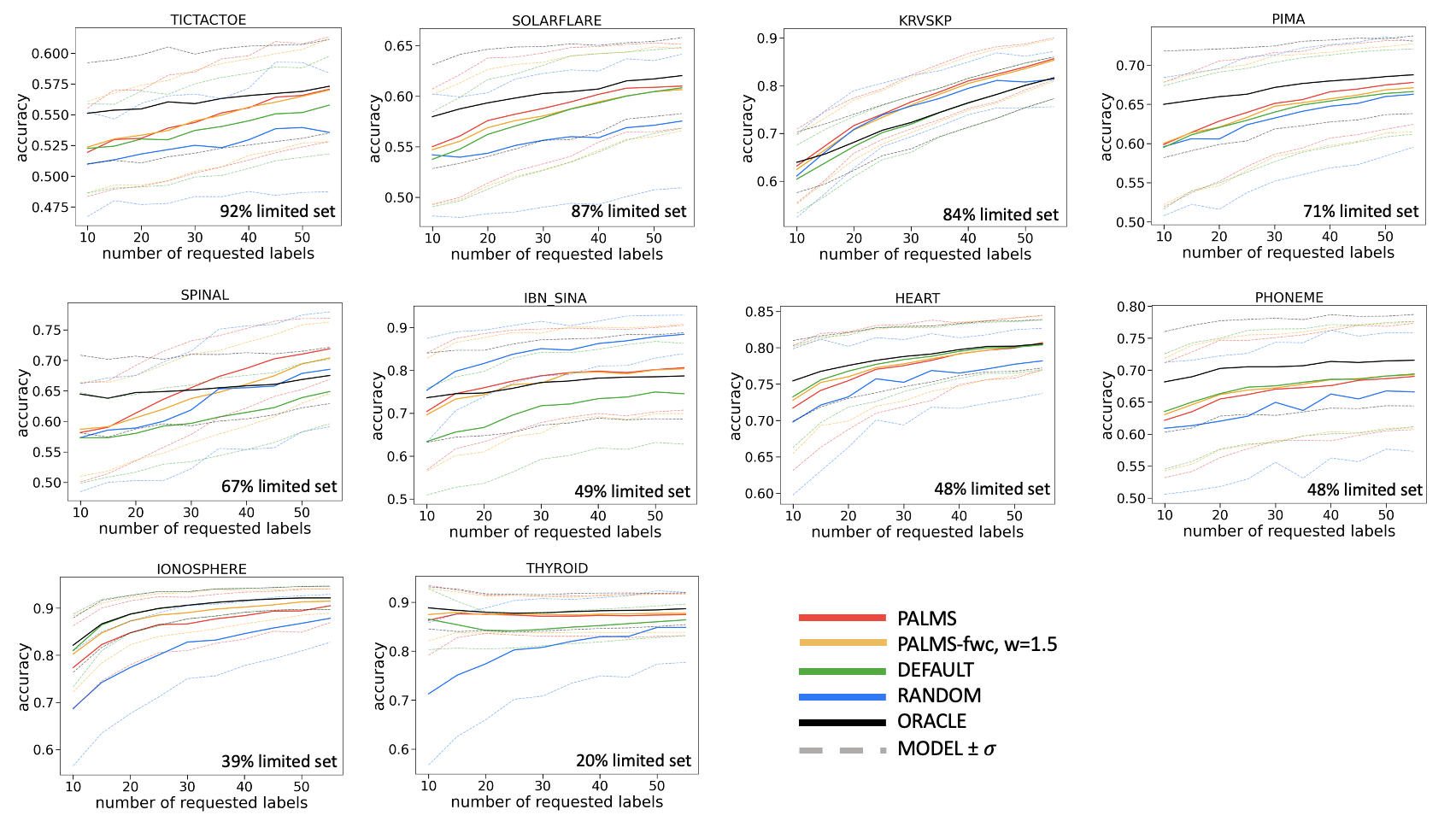}
    \caption{Comparisons of test accuracy for binary classification on five different datasets. The PALMS method achieves the closest performance to ORACLE for the Tictactoe, Solarflare, KRvsKP, Pima, Spinal, and Thyroid datasets. Note that except for the Thyroid dataset, these datasets all have a high percentage of datapoints in the limited set (see Table~\ref{tab:datasets}). For the IBN\_Sina dataset, the PALMS and PALMS-fwc methods both achieve better performance than DEFAULT and eventually even outperform ORACLE. The DEFAULT model achieves closer performance to ORACLE for the Heart, Phoneme, and Ionosphere datasets. For the Ionosphere dataset, the DEFAULT model is often the ORACLE model. Both datasets have a lower percentage of datapoints in the limited set. Note that when this is the case, the overall performance of the model is higher since a smaller limited set generally corresponds to an easier binary classification problem. Thus, even though PALMS does not perform as well as DEFAULT in these cases, it is still performing acceptably well. Finally, we see that PALMS-fwc achieves a tradeoff between improving performance on the datasets with smaller limited sets (Ionosphere, Heart, Phoneme, Thyroid), and slightly decreasing performance on the datasets with larger limited sets (Tictactoe, Solarflare, KRvsKP, Pima, Spinal, IBN\_Sina). Note that we typically do not know \textit{a priori} which type of dataset we have for a given problem, so PALMS-fwc may be the optimal choice unless domain-specific information is available.}
    \label{fig:performance}
\end{figure*}

In Figure~\ref{fig:performance}, we show the mean accuracy of each method on the test set (solid lines) as well as the corresponding standard deviations $\sigma$ (dashed lines). For the Tictactoe, Solarflare, KRvsKP, Pima, and Spinal datasets, we see that PALMS and PALMS-fwc both outperform RANDOM and DEFAULT, achieving the closest performance to ORACLE. Note that these datasets all have a high percentage of datapoints in the limited set, which means that they are likely difficult to classify. This is corroborated by the maximum accuracies achieved after 55 requested labels (59 total labeled datapoints, including the 4 initial samples). PALMS and PALMS-fwc also achieve top performance on the Thyroid dataset, which is so easily separable that all but the RANDOM model were able to achieve over 85\% accuracy with only 10 datapoints. For the IBN\_Sina dataset, the PALMS and PALMS-fwc methods both achieve better performance than DEFAULT and eventually even outperform ORACLE. It is possible to outperform ORACLE because ORACLE relies on a single fixed model to actively sample datapoints and train the final classifier, whereas PALMS and PALMS-fwc may select a different model to train the final classifier from the default fixed model used to actively sample datapoints. However, we note in this case that RANDOM outperforms all methods; this is unfortunately also possible due to the biased sampling in active learning, which may have difficulty correcting from poor initial models that may arise with complex data distributions~\cite{Attenberg2011}.

For the Heart, Ionosphere, and Phoneme datasets, the DEFAULT model achieves closer performance to ORACLE. In fact, for the Ionosphere dataset, the DEFAULT model is often equivalent to the ORACLE model. However, we note that these three datasets have a lower percentage of datapoints in the limited set, corresponding to easier classification and higher achieved accuracies. Thus, even though PALMS does not perform as well as DEFAULT in these cases, it is still achieving acceptable performance if we had decided to use PALMS rather than DEFAULT in a real use case scenario. 

Finally, we see that PALMS-fwc with a weight parameter of $w=1.5$ achieves a tradeoff between improving performance on the datasets with smaller limited sets (Ionosphere, Heart, Phoneme, Thyroid), and slightly decreasing performance on the datasets with larger limited sets (Tictactoe, Solarflare, KRvsKP, Pima, Spinal, IBN\_Sina). Since we typically do not know \textit{a priori} which type of dataset we have for a given problem, PALMS-fwc may be the optimal choice unless domain-specific information is available.

\section{Conclusions}
\label{sec:conclusions}

In this work, we proposed two new approaches called PALMS and PALMS-fwc for active learning with model selection that is especially suitable for real world problems where the labeling budget might be extremely small, up to a few dozen datapoints. Using pool-based active learning with SVC using a RBF kernel as the underlying active learning model, we show with PALMS that model selection for the optimal hyperparameters $(C^*, \gamma^*)$ can be implemented as an additional step after active learning by reusing the actively sampled data for leave-one-out cross validation (LOOCV). Even though this actively sampled dataset is inherently biased, we show empirically that it is still useful for model selection, especially on difficult to classify datasets. 

Furthermore, we demonstrate that one weakness of LOOCV on the actively sampled data is that the LOOCV error for the underlying fixed model used for active learning underestimates performance accuracy on the test set, especially for datasets that are easier to classify, since by construction there are proportionally more actively sampled datapoints in the uncertain regions close to the decision boundary compared to in the more certain regions away from the decision boundary. Therefore, we developed PALMS-fwc to correct for this by weighting the actively sampled datapoints when computing LOOCV scores for the fixed model. We see empirically that for datasets that are easier to classify, this correction improves model selection performance; however the tradeoff is that for datasets that are more difficult to classify, the underlying assumption of how the actively sampled dataset is biased may not be as correct, and model selection performance is slightly degraded. Since we cannot know \textit{a priori} what kind of dataset we have, PALMS-fwc with a small weight is recommended in general. When there is prior information about the dataset, we may instead choose to use PALMS (when the dataset is likely to be difficult to classify) or PALMS-fwc with a larger weight (when the dataset is likely to be easy to classify).

Our method does not require a separate labeled validation set, which would need to be sufficiently large in order to effectively distinguish between different models, and thus is not a feasible approach when the total labeling budget is small. It also does not try to incorporate model selection in the active learning process when acquiring new labeled datapoints. Since our estimates of model performance will have high variance especially in the beginning of active learning when the labeled dataset is very small, it is not desirable to use these unreliable estimates during active sampling since it may degrade the performance of the active learner. Our method avoids this issue by implementing model selection after active learning on a fixed model is complete.

In future work, we will explore the effect of varying the underlying fixed model that is used to generate the actively sampled labeled data for model selection, as well as more advanced methods for determining the optimal weight correction to use in PALMS-fwc. We will also consider classifiers other than SVC, problems with imbalanced classes, and multi-class classification. Finally, since our actively sampled datasets are so small, we will explore methods to prevent overfitting of the model selection process when there are a comparatively large number of models to choose from.

%%% References Outline
\nocite{Cohn1996}
\nocite{Settles2012}
\nocite{Settles2011}
\nocite{Attenberg2011}
\nocite{Lowell2019}
\nocite{Chapelle1999}
\nocite{Chapelle2002}
\nocite{Cherkassky2003}
\nocite{Gardner2015}
\nocite{Malkomes2016}
\nocite{Malkomes2018}
\nocite{Campbell2000}
\nocite{Schohn2000}
\nocite{Tong2001}
\nocite{Kremer2014}
\nocite{Gress2015}
\nocite{Kottke2017}
\nocite{Bach2006}
\nocite{Baram2004}
\nocite{Sugiyama2005}
\nocite{Sugiyama2008}
\nocite{Ali2014}
\nocite{Dasgupta2011}
\nocite{Baldridge2004}
\nocite{Tomanek2007}
\nocite{Tomanek2011}
\nocite{Hu2016}
\nocite{Sugiyama2007}
\nocite{Beygelzimer2009}
\nocite{Tang2019}

% references section
\bibliographystyle{IEEEtran}
\bibliography{palms}

\end{document}